\documentclass[letterpaper, 10pt, conference]{ieeeconf}

\IEEEoverridecommandlockouts
\usepackage{algorithm}
\usepackage{algorithmic}
\usepackage{amsmath}
\usepackage{dsfont}
\usepackage{graphicx}
\usepackage{xcolor}
\usepackage[caption=false]{subfig}
\usepackage[colorlinks]{hyperref}

\def\x{{\times}}
\def\eg{\emph{e.g.}}

\def\etal{\emph{et al.}}
\def\Eqref#1{Eq.~\ref{#1}}

\newcommand{\app}{\raise.17ex\hbox{$\scriptstyle\sim$}}
\newcommand{\codecomment}[1]{{\color{gray}{#1}}}
\newcommand{\includegraphicsgenim}[2]{\includegraphics[trim=165 40 165 165, clip, #1]{#2}}
\newcommand{\includegraphicstriml}[3]{\includegraphics[trim=#1 0 0 0, clip, #2]{#3}}
\renewcommand{\paragraph}[1]{\textbf{#1}}

\begin{document}
\title{\LARGE \bf
State-Only Imitation Learning for Dexterous Manipulation
}
\author{Ilija Radosavovic$^{1}$, Xiaolong Wang$^{2}$, Lerrel Pinto$^{3}$, Jitendra Malik$^{1}$%
\thanks{*Supported by DARPA program on Machine Common Sense}%
\thanks{$^{1}$UC Berkeley $^{2}$UC San Diego $^{3}$New York University}%
}
\maketitle
\thispagestyle{empty}
\pagestyle{empty}

\begin{figure}[t]
\begin{minipage}{1.0\textwidth}\centering
\includegraphics[width=0.85\textwidth]{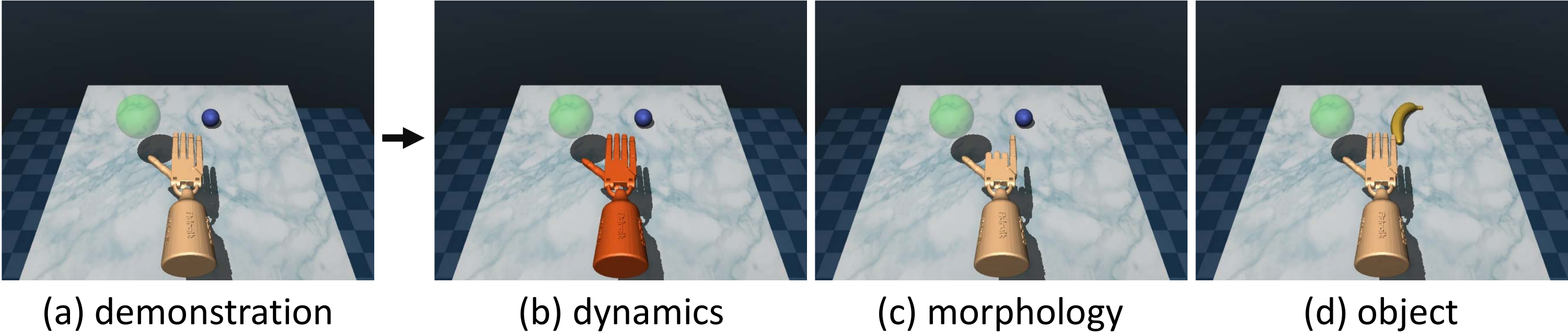}
\caption{\textbf{Problem.} We explore \emph{state-only} imitation learning for dexterous manipulation. Compared to standard state-action imitation learning, this regime has the potential to enable learning from demonstrations coming from different but related settings. We show that using our method, called \emph{SOIL}, we can effectively leverage demonstrations of relocating a ball to a target position (a), to learn to perform the same task with different dynamics (b), morphologies (c), and objects (d). This relaxed problem setting brings us closer to the more general setting of learning from internet videos of humans.}
\label{fig:teaser}
\end{minipage}\vspace{-4mm}
\end{figure}

\begin{abstract}
Modern model-free reinforcement learning methods have recently demonstrated impressive results on a number of problems. However, complex domains like dexterous manipulation remain a challenge due to the high sample complexity. To address this, current approaches employ expert demonstrations in the form of state-action pairs, which are difficult to obtain for real-world settings such as learning from videos. In this paper, we move toward a more realistic setting and explore state-only imitation learning. To tackle this setting, we train an inverse dynamics model and use it to predict actions for state-only demonstrations. The inverse dynamics model and the policy are trained jointly. Our method performs on par with state-action approaches and considerably outperforms RL alone. By not relying on expert actions, we are able to learn from demonstrations with different dynamics, morphologies, and objects. Videos available on the \href{https://people.eecs.berkeley.edu/~ilija/soil}{project page}.
\end{abstract}

\section{INTRODUCTION}

Dexterous manipulation with multifingered hands has the potential to equip robots with human-like dexterity and enable them to generalize across diverse environments, goals, and tools. However, it comes at a significant cost: complex high-dimensional action spaces. Traditionally, this has been a challenge for standard model-based trajectory optimization approaches. Recently, there has been a renewed interest in using reinforcement learning (RL) techniques for dexterous manipulation. For example, we have witnessed impressive results in using large-scale RL with domain randomization for dexterous in-hand manipulation tasks~\cite{Openai2018, Openai2019}.

In current instantiations, such techniques suffer from high sample complexity and require considerable human effort in reward function design. These issues are even more severe in the case of dexterous manipulation with multifingered hands. In particular, high-dimensional action spaces make exploration and optimization challenging, and it is hard to specify effective reward functions for dexterous tasks.\\
\\

\vspace{42.6mm}
To enable sample-efficient learning of complex tasks, one promising avenue lies in imitation learning or learning from demonstrations~\cite{Bakker1996, Schaal1999}. Learning by imitation is a powerful mechanism in the cognitive development of children~\cite{Tomasello1993, Meltzoff1995}. Similar to learning in children, our robots could acquire motor skills by learning from demonstrations. This general paradigm also has the potential to enable learning from internet-scale videos of humans in the wild.

Overall, there has already been a considerable progress toward this goal. In particular, a large body of work has focused on the setting with demonstrations in the form of state-action pairs. For example, Rajeswaran \etal~\cite{Rajeswaran2018} collect demonstrations for dexterous manipulation using a virtual reality headset and a motion capture glove. They show that augmenting the RL objective with an imitation term can lead to large gains in sample complexity compared to RL alone. However, collecting data in such setups is time-consuming and limited to only a subset of tasks. Moreover, requiring actions makes it impossible to leverage the readily available internet videos of humans as demonstrations.

Relying on state-action demonstrations raises another challenge: learning in the presence of \emph{demonstration mismatch}. In particular, learning a policy that is constrained to output the same action as the demonstrated state-action pair limits the flexibility of the policy and hinders the ability to leverage demonstrations from different settings (Figure~\ref{fig:teaser}). Is relying on demonstrator actions necessary? Can we extract useful information from state-only demonstrations?

In this paper, we move toward the more general setting and explore learning from demonstrations containing only states without the actions. To tackle this setting, we propose a simple and effective method that we call SOIL (for \emph{State-Only Imitation Learning}). Our method involves an agent that has a behavior policy and an inverse dynamics model. The agent uses the policy to interact with the world and learns an inverse model from past interactions by self-supervision. Using the inverse model, the agent predicts the actions that best achieve the observed state-only demonstrations. We train the policy to optimize the task reward using both RL and demonstrations with predicted actions. The agent can thus rely on demonstrations in early stages of training and learn to surpass them later on. The policy and the inverse dynamics model are trained jointly in an alternating fashion.

We evaluate our method in simulation on four different dexterous manipulation tasks~\cite{Rajeswaran2018}. Interestingly, we find that SOIL performs on par with state-action approaches. Moreover, our method achieves considerable improvements compared to RL without demonstrations. We further show that SOIL can leverage demonstrations with different dynamics, morphologies, and objects (Figure~\ref{fig:teaser}), while using the demonstrator actions degrades the performance. Our results suggest that relying on state-action demonstrations may not be necessary for developing imitation learning methods.

\paragraph{Contributions:} (1) We explore state-only imitation learning for dexterous manipulation. This regime reduces the burden of data collection, allows learning with demonstration mismatch, and brings us closer to learning from videos. (2) We propose SOIL, a simple and effective method for state-only imitation learning. The core idea is to train a policy and an inverse model jointly. The policy is trained using RL and demonstrations with predicted actions, and the inverse model is trained using self-supervision. (3) We show that our method can perform on par and even outperform state-action approaches when learning with demonstrations mismatch.

\section{RELATED WORK}

\paragraph{Dexterous manipulation.} There is a wide range of work on dexterous manipulation with optimization and planning~\cite{Dogar2010, Bai2014, Andrews2013}. However, these approaches are often insufficient for solving complex tasks. Recently, we witnessed promising results achieved by RL in simulation~\cite{Openai2018, Openai2019}. These policies can be deployed to the real robot with domain randomization~\cite{Sadeghi2017, Tobin2017}. However, this requires large-scale training and is hard to generalize to new environments.

\paragraph{Imitation learning with BC.} One way to overcome these challenges is to perform imitation learning with expert demonstrations. A common approach to imitation learning is behavior cloning~\cite{Pomerleau1989, Bain1995, Bojarski2016}. BC has two limitations: it requires expert actions and the learnt policy is upper-bounded by expert performance. Similar to ours, \cite{Torabi2018b} infer actions by training an inverse model together with the policy. However, their BC policy is limited by the expert performance. In contrast, our approach that uses the RL objective is not.

\paragraph{RL with demonstrations.} To inherit benefits from both RL and imitation learning, researchers explored RL with demonstrations~\cite{Peters2008, Duan2016, Vevcerik2017, Peng2018d}. For example,~\cite{Vevcerik2017} propose to use demonstrations with off-policy methods by adding them to the replay buffer. While off-policy methods can be more sample efficient, they are generally less stable and scale worse to high-dimensional spaces~\cite{Rajeswaran2018, Duan2016}. Instead of using off-policy methods, we incorporate demonstrations with on-policy methods via an auxiliary term. Furthermore, our expert demonstrations contain only states without actions. Thus, previous approaches cannot be adopted directly.

\paragraph{Inverse RL.} If we do not have access to actions, one option would be to learn a density model akin to inverse reinforcement learning~\cite{Russell1998, Ng2000, Abbeel2004, Fu2017, Aytar2018, Ho2016, Zhu2018, Torabi2018g, Sun2019, Liu2020}. For example,~\cite{Ho2016} learn a density model implicitly to estimate if a trajectory is from demonstrations or the policy, and use the output as the reward in an adversarial training framework. However, learning density models for high dimensional tasks like dexterous manipulation is often inefficient, and the adversarial objectives often collapse into sub-optimal modes which makes learning unstable.

\paragraph{Learning dynamics models.} One way to improve the sample efficiency of RL is to learn the dynamics model. For example,~\cite{Nagabandi2019} showed that model-based RL could provide an efficient way of learning dexterous manipulation tasks. Besides training forward models, researchers have also explored learning inverse models~\cite{Kawato1987, Pinto2016, Agrawal2016, Christiano2016, Nair2017, Pathak2017, Pathak2018, Edwards2018, Torabi2018b, Kumar2019, Pavse2020}. The inverse model is typically pretrained and then used to complete expert demonstrations. In contrast, we train the inverse dynamics model jointly with the policy that optimizes the task reward. Most similar to our approach is~\cite{Guo2019} that focuses on games with discrete actions. Instead, we focus on complex dexterous manipulation tasks with high-dimensional continuous action spaces.

\paragraph{Following demonstrations.} Another line of work in imitation learning is to train policies to follow expert demonstrations~\cite{Liu2017, Peng2018s, Sharma2018, Sermanet2018}. For example,~\cite{Peng2018s} show successful imitation of human motion from videos. However, the policy is only trained for repeating one particular trajectory. Instead of following an expert trajectory, our policy is goal-conditioned and optimizes the task reward.

\section{PRELIMINARIES}

We begin by briefly discussing the relevant RL and imitation methods we build upon in this work.

\paragraph{Reinforcement learning.} We focus on the policy gradient methods that directly maximize the expected sum of discounted rewards using gradient ascent. In its simplest form, the vanila policy gradient~\cite{Williams1992} is given by:
\begin{align}
g = \sum_{(s, a) \in \pi} \nabla_\theta \log \pi_\theta(a|s) A^\pi(s,a),
\label{eq:g}
\end{align}
where $A^\pi(s, a)$ is the advantage function. In all experiments, we use the more effective natural policy gradient (NPG)~\cite{Kakade2002}. However, it is still hard to obtain good performance due to exploration and optimization challenges, especially in complex domains like dexterous manipulation.

\paragraph{Imitation learning.} One way to overcome these challenges is to employ demonstrations in the form of state-action pairs. To utilize state-action demonstrations,~\cite{Rajeswaran2018} propose demo augmented policy gradient (DAPG), which augments the vanilla policy gradient with an auxiliary imitation term:
\begin{align}
\label{eq:dapg}
g_{dapg} =  g + \lambda_{0} \lambda_{1}^{k}  \sum_{(s, a) \in D} \nabla_\theta \log \pi_\theta(a|s) .
\end{align}
where $D$ are the state-action demonstrations, and $\lambda_0$ and $\lambda_1$ are scalar hyperparameters that control the relative contribution of the imitation learning objective.

\newpage
\section{STATE-ONLY IMITATION LEARNING}

\begin{figure}\centering
\includegraphics[width=0.85\linewidth]{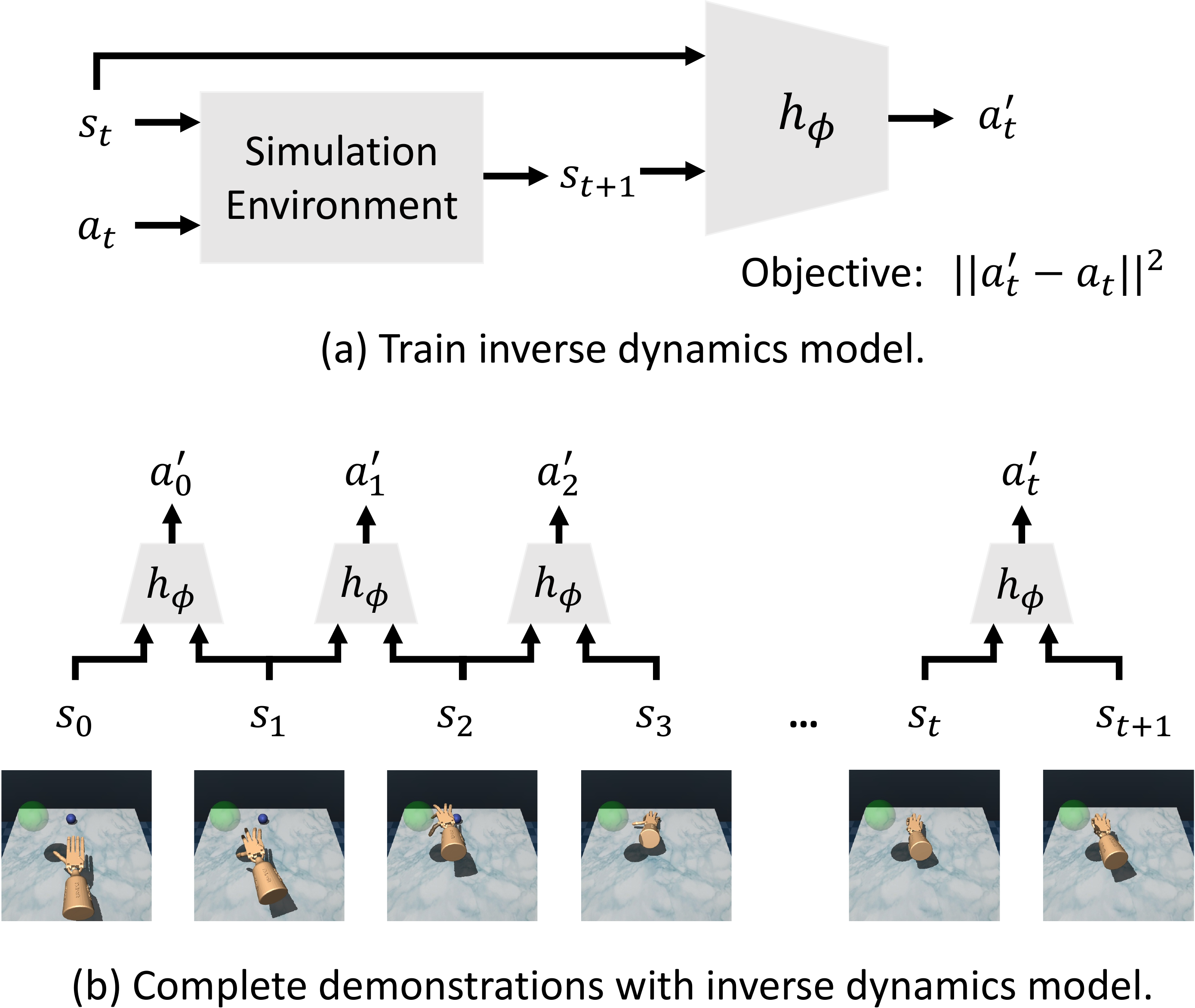}\vspace{-0mm}
\caption{\textbf{Method.} We present \emph{SOIL}, a simple and effective method for \emph{state-only} imitation learning. The basic idea is to train a policy and an inverse model jointly. To train the inverse model, we use the policy to generate trajectories from the environment and perform self-supervised learning (a). Using the inverse model, we predict actions to complete the state-only demonstrations (b). The details of the method are given in Algorithm~\ref{alg:soil}.}
\label{fig:model}\vspace{-1mm}
\end{figure}

Here we describe the problem setting and our approach.

\subsection{Problem Setup}

In this work, we focus on a more realistic setting for imitation learning, where demonstrations only contain states and no actions. Leveraging state-only demonstration allows for practical imitation learning since state-estimation from expert demonstrators is more feasible. One example is learning from videos of human demonstrations, where state estimates are readily available~\cite{Handa2019} while human action estimates are not. Hence, this relaxed problem setting brings us closer to real-world settings and potentially utilizing videos of humans.

As a first step toward this goal, we assume our model has direct access to the states represented by the joint angles of the hand, forces applied to the joints, and the speed of the joints. We also provide the locations of the objects. This allows us to investigate the effects of state-only imitation without conflating with state estimation. Our policy is trained in the same state space as the demonstrations.

We note that even if the state space is well aligned, our problem is still challenging without access to the actions between every two states. Recall that both behavior cloning and DAPG mentioned in the previous section require the ``ground-truth'' actions. Thus, we cannot directly apply standard imitation learning objectives for training our policy. Can we predict the actions from the provided state-only demonstrations? If we can predict these actions well, we can stand on the shoulders of powerful behavior cloning techniques and achieve good performance. Our learning framework revolves around this idea of action prediction.

Concretely, we train an inverse dynamics model, which takes two consecutive demonstration states as inputs and estimates the action as the output. A naive way to collect the training data for supervising this inverse dynamics model is to sample random trajectories. However, the action space is too large to explore randomly with a robot hand. To acquire informative training data for the inverse model, we need a reasonable policy to explore the action space. On the other hand, to train a reasonable manipulation policy, we need to predict the correct actions with the inverse model.

To overcome this problem, we propose to train the inverse dynamics model and the policy network jointly in an iterative manner. Thus, the policy network can help the inverse model better explore the action space, and a better inverse model, in turn, provides better training examples for the policy network. In the following, we describe the inverse model and our learning framework for joint iterative training.

\begin{algorithm}[t]
\caption{State-Only Imitation Learning (SOIL)}
\label{alg:soil}
\begin{algorithmic}
  \STATE {\bfseries Input:} Inverse model $h$, Policy $\pi$, Replay buffer $R$,\\State-Only Demonstration $D$.
  \STATE {\bfseries Initialize:} Learnable parameters $\phi$ for $h_\phi$, $\theta$ for $\pi_\theta$.
  \FOR {$i$=1,2,..$N_\text{iter}$}
  \STATE \codecomment{\# Collect trajectories}
  \STATE $\tau_i \equiv \{s_t, a_t, s_{t+1}, r_t \}_i \sim \pi_{\theta}$
  \STATE \codecomment{\# Add data to the buffer}
  \STATE $R \leftarrow R \cup \tau_i$
  \FOR {$j$=1,2,..$N_\text{inv}$}
  \STATE \codecomment{\# Sample a batch of state-action tuples}
  \STATE $B_j \equiv \{s_t, a_t, s_{t+1}\}_j \sim R$
  \STATE \codecomment{\# Update the inverse dynamics model}
  \STATE $\phi \leftarrow \text{InvOpt}(B_j ; \phi)$, according to \Eqref{eq:inf:inv}
  \ENDFOR
  \STATE \codecomment{\# Predict actions using the inverse model}
  \STATE $D^{\prime} \leftarrow$ complete $D$ with  $h_\phi$, according to \Eqref{eq:inf:inv}.
  \STATE \codecomment{\# Perform SOIL policy gradient update}
  \STATE $\theta \leftarrow \text{PolicyOpt}(\tau_i, D^{\prime}; \theta)$, according to \Eqref{eq:soil}
  \ENDFOR
\end{algorithmic}
\end{algorithm}

\begin{figure*}[t]\centering
\includegraphics[width=0.9\linewidth]{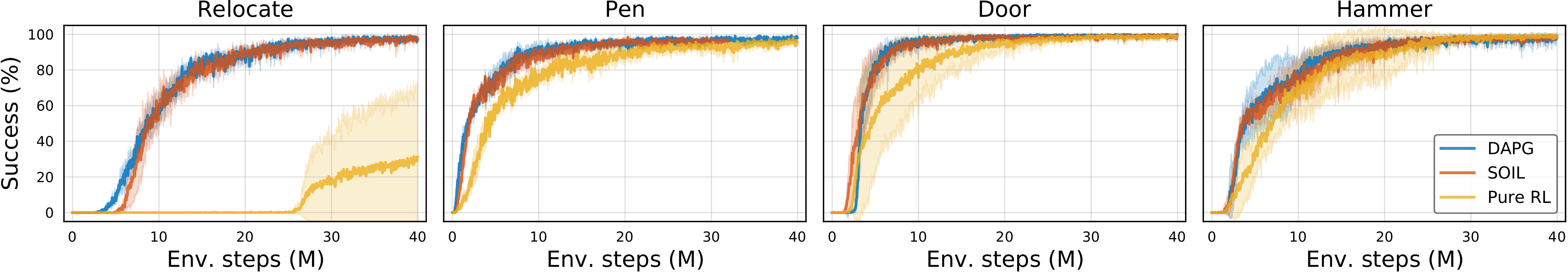}\vspace{-1mm}
\caption{\textbf{State-action comparisons.} We perform controlled comparisons of our state-only method SOIL to DAPG that uses state-action demonstrations \emph{(upper-bound)} and pure RL without demonstrations \emph{(lower-bound)}. We observe consistent results across four different dexterous manipulation tasks: SOIL performs comparably to DAPG while being considerably better than pure RL. Note that the gap compared to pure RL is the largest for the most challenging object relocation task (while perhaps unintuitive at first, object relocation is considerably harder than the remaining tasks due to exploration challenges).}
\label{fig:comparison_state_action}\vspace{-0mm}
\end{figure*}

\begin{figure*}[t]\centering
\includegraphics[width=0.9\linewidth]{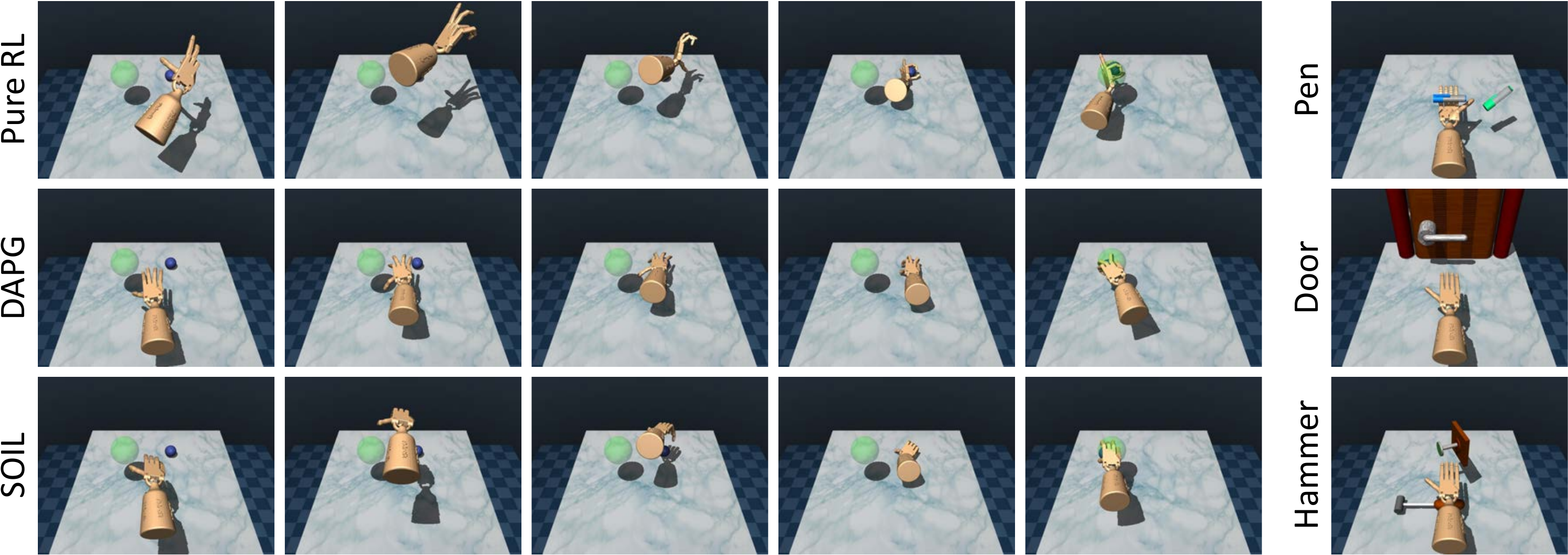}\vspace{-1mm}
\caption{\textbf{Qualitative examples.} \emph{Left:} We show selected episodes for the object relocation task. We see that our SOIL policy (bottom) is comparable to the state-action DAPG policy (middle) while being considerably more realistic than the pure RL policy (top). \emph{Right:} We observe the consistent general trends for the remaining three tasks: in-hand pen manipulation, door opening, and tool use. We encourage the readers to check the \href{https://people.eecs.berkeley.edu/~ilija/soil}{project page} for videos.}
\label{fig:qualitative}\vspace{-2mm}
\end{figure*}

\subsection{Inverse Dynamics Model}

We propose to train an inverse dynamics model which takes two consecutive states as inputs and predicts the action between them. As shown in Figure~\ref{fig:model}a, we assume that we have the state and action triplet of the form $(s_t, a_t, s_{t+1})$. Our inverse dynamics model is a small MLP network $h_\phi$ which is parameterized by $\phi$. We can then use $h_\phi$ to predict the action $a_t^{\prime}$ given two states (\Eqref{eq:inf:inv}).

We train the inverse model in a self-supervised learning manner using the L2 loss. Given a policy network, we sample trajectories from the policy to generate the training data for training the inverse model. In particular, we collect the state and action triplets and add them to a replay buffer $R$. We then sample uniformly from $R$ and obtain a batch of training examples $B$. The training objective is computed following \Eqref{eq:inf:inv} and averaged over the batch $B$.
\begin{align}
\label{eq:inf:inv}
a_t^{\prime} = h_\phi(s_t, s_{t+1}) , \quad L_\text{inv} = ||a_t^{\prime} - a_t||^2 .
\end{align}
Learning an inverse model is appealing. First, it is adaptable which enables utilizing demonstrations from different settings (\eg, different dynamics). Second, it is agent-specific and may thus generalize to new settings (even if the task the agent faces is new, its hand still works the same way).

\subsection{Joint Training Procedure}

We propose to jointly train the inverse dynamics model and the policy network. Since the actions are not directly provided from the demonstrations but estimated by the inverse dynamics model, we adjust the DAPG objective (\Eqref{eq:dapg}). As shown in Figure~\ref{fig:model}b, given the states from the demonstration, we use the inverse model to predict the actions $a^\prime$ between the consecutive states (Eq.~\ref{eq:inf:inv}), and generate the new demonstration set with state and action pairs $D^\prime$. We incorporate the predicted actions into the overall policy gradient as,
\begin{align}
\label{eq:soil}
g_{soil} =  g + \lambda_{0} \lambda_{1}^{k}  \sum_{(s, a^{\prime}) \in D^{\prime}} \nabla_\theta \log \pi_\theta(a^{\prime}|s).
\end{align}
The expression consists of a policy gradient term and an auxiliary imitation term. We anneal the imitation term to zero by setting $\lambda_{1} < 1$ and increasing $k$ with the number of training iterations. Intuitively, we rely on demonstrations to explore the space initially and gradually transition to policy gradient which has the potential to surpass demonstrations in later stages of training. During training, we jointly optimize the objective for the inverse dynamics model (Eq.~\ref{eq:inf:inv}) and the objective for the policy network (Eq.~\ref{eq:soil}). The details of the joint optimization process are given in Algorithm~\ref{alg:soil}.

\section{EXPERIMENTS}

We now evaluate our method in simulation. We perform controlled comparisons to state-action and state-only methods. We further present ablations and study robustness to demonstration mismatch.

\subsection{Experimental Setup}

\paragraph{Simulator.} We use the MuJoCo physics simulator tailored to robotics-related tasks~\cite{Todorov2012}. We adopt the simulated model of the dexterous hand provided in the ADROIT platform~\cite{Kumar2013}, designed for dexterous manipulation tasks. The hand model has five fingers and 24 degrees of freedom, which involve position control and joint angle sensors.

\paragraph{Tasks.} The suite proposed in~\cite{Rajeswaran2018} consists of four dexterous manipulation tasks: object relocation (pick up the ball and take it to the target position), in-hand manipulation (position the pen to match the desired orientation), door opening (undo the latch and open the door), and tool use (pick up the hammer and drive the nail into the board). See Figure~\ref{fig:qualitative}.

\paragraph{Demonstrations.} We use the demonstrations collected by~\cite{Rajeswaran2018} using a virtual reality headset and a motion capture glove~\cite{Kumar2015}. There are 25 demonstrations per task. Each demonstration is a sequence of state-action pairs. For state-only methods, we ignore the provided actions.

\paragraph{RL algorithm.} NPG is the current state-of-the-art RL algorithm for these tasks~\cite{Rajeswaran2018} and we use it for all methods in our experiments. In addition, we verified that our method can be adapted to PPO~\cite{Schulman2017} as well (not shown).

\begin{figure}[t]\centering
\includegraphics[width=0.9\linewidth]{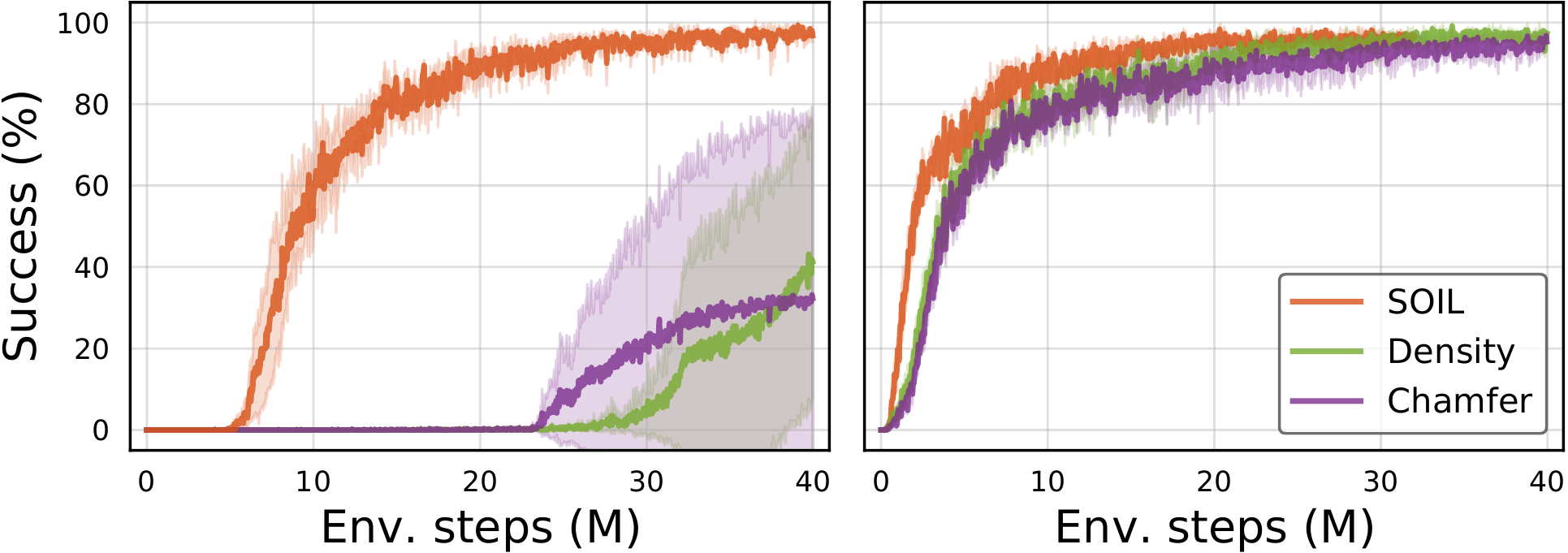}\vspace{-1mm}
\caption{\textbf{State-only comparisons.} We compare SOIL to two state-only baselines: state matching with Chamfer distance and density estimation. Our method outperforms the baselines on both the more challenging object relocation task (left) and the pen in-hand manipulation task (right).}
\label{fig:comparison_state_only}\vspace{-0mm}
\end{figure}

\begin{figure}[t]\centering
\includegraphics[width=0.9\linewidth]{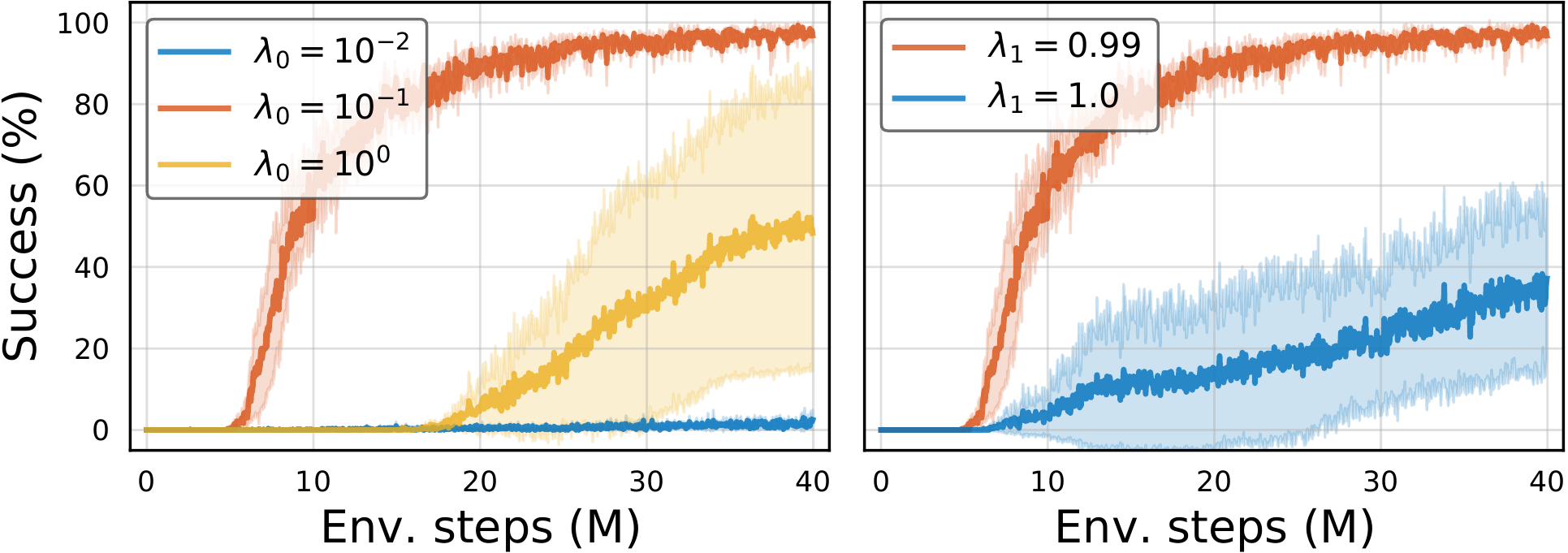}\vspace{-1mm}
\caption{\textbf{Imitation term weights.} We study the impact of the imitation term weights $\lambda_0$ and $\lambda_1$ from \Eqref{eq:soil}. \emph{Left:} The performance is sensitive to the scale of the imitation term. \emph{Right:} Annealing the imitation term to zero ($\lambda_1 = 0.99$) is beneficial. \emph{Task:} Relocate.}
\label{fig:ablations_lams}\vspace{-2mm}
\end{figure}

\subsection{State-Action Comparisons}\label{sec:state_action}

We first evaluate our state-only method in the standard imitation learning setup. We compare with two methods: DAPG that uses state-action demonstrations and pure RL without demonstrations (NPG).

\paragraph{Pure RL comparisons.} We consider pure RL as a lower-bound for our approach. Intuitively, using demonstrations---even without actions---should not hurt the performance and should work at least as well as pure RL. In Figure~\ref{fig:comparison_state_action}, we compare our state-only approach to pure RL on four different tasks. In all cases, we observe that our method SOIL outperforms pure RL. We highlight two cases next. First, the gap is the largest for the hardest task of object relocation (left). This is a promising signal for the applicability of our method to more challenging settings. Second, pure RL works surprisingly well on the hammer task (right).

\paragraph{DAPG comparisons.} Similarly, we consider DAPG that uses state-action demonstrations as an upper-bound for our approach because our inverse model trained from scratch is unlikely to predict actions superior to the ground-truth actions. In Figure~\ref{fig:comparison_state_action}, we compare our method SOIL to DAPG on four different tasks. In all cases, the findings are again consistent: SOIL comes surprisingly close to DAPG that uses ground-truth actions. This is very encouraging and suggests that having access to actions may not be critical for developing well-performing imitation learning approaches.

\paragraph{Qualitative examples.} The returns alone do not paint a complete picture. To get a better sense of the learnt policies, we show selected episodes in Figure~\ref{fig:qualitative} (left). First, we observe that although the pure RL policy can manage to complete the task it does not exhibit realistic behavior (top). In contrast, the DAPG policy is much more realistic (middle). Lastly, we see that our SOIL policy is comparable to DAPG while being more realistic than pure RL (bottom).

\paragraph{Summary.} We evaluate our method in a controlled setting where state-action demonstrations are available. We find that our state-only method reaches the performance of state-action DAPG while considerably outperforming RL without demonstrations. Overall, our results suggest that relying on expert actions may not be necessary for imitation learning.

\begin{figure}[t]\centering
\includegraphics[width=0.9\linewidth]{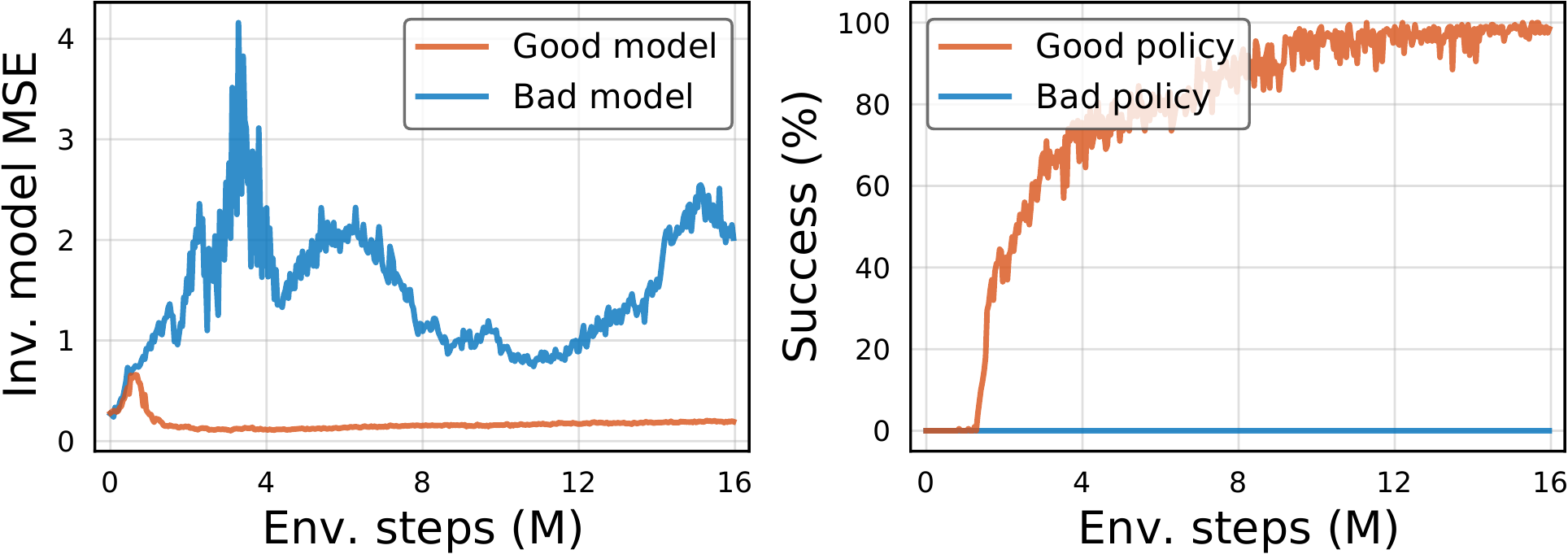}\vspace{-1mm}
\caption{\textbf{Inverse dynamics model.} We study the impact of the quality of the inverse model (left) on the policy (right). We observe a clear correlation between the two: a good inverse model (low error) results in a good policy (high return). \emph{Task:} Hammer.}
\label{fig:ablations_model}\vspace{-0mm}
\end{figure}

\begin{figure}[t]\centering
\includegraphics[width=0.9\linewidth]{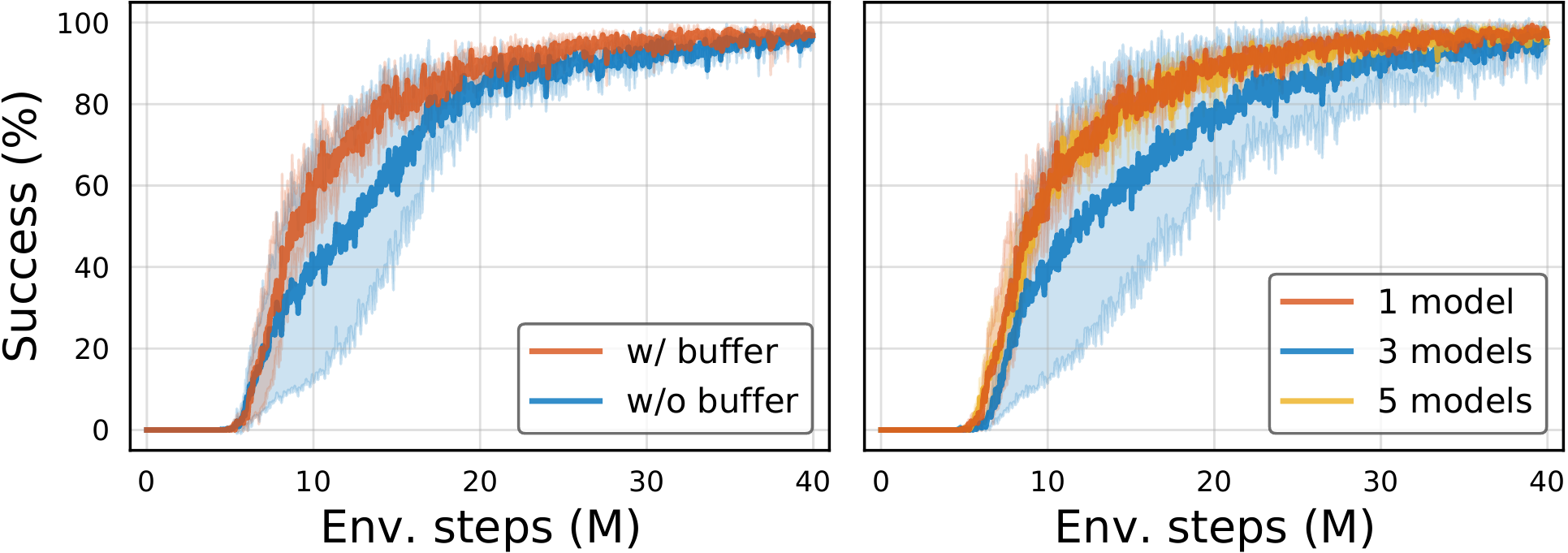}\vspace{-1mm}
  \caption{\textbf{Model-based RL enhancements.} \emph{Left:} Using a replay buffer to aggregate the training data for the inverse dynamics model reduces the variance. \emph{Right:} Using an ensemble of inverse dynamics models does not improve the performance. \emph{Task:} Relocate.}
\label{fig:ablations_tricks}\vspace{-2mm}
\end{figure}

\begin{figure*}[t]\centering
\begin{minipage}{.33\linewidth}\centering
 \subfloat[\textbf{Dynamics.}\label{fig:gen_dynamics_mass}]{\shortstack{
 \includegraphics[height=21.5mm]{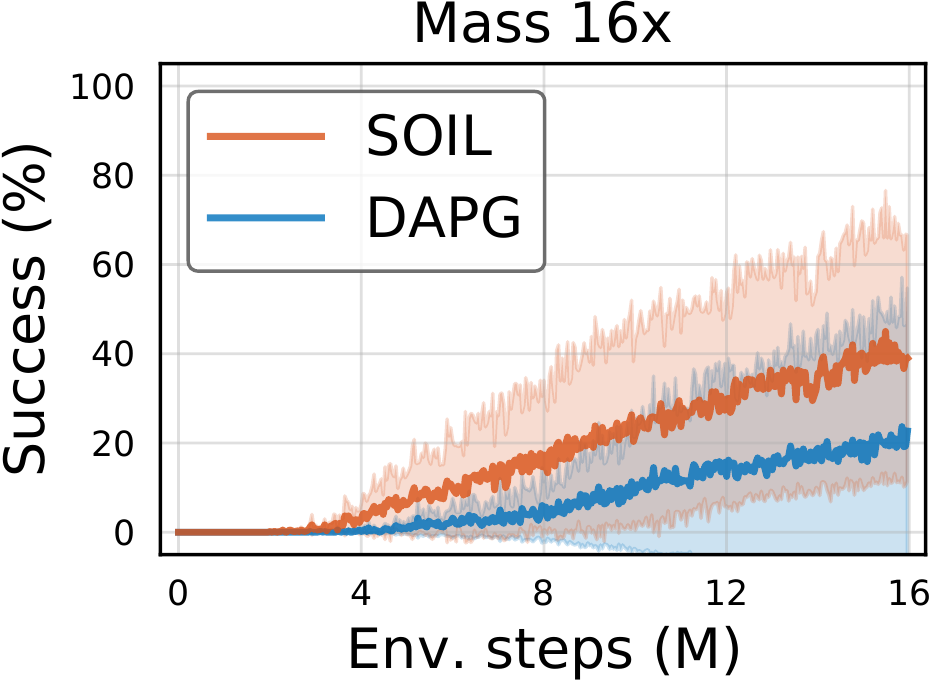}
 \includegraphicstriml{44}{height=21.5mm}{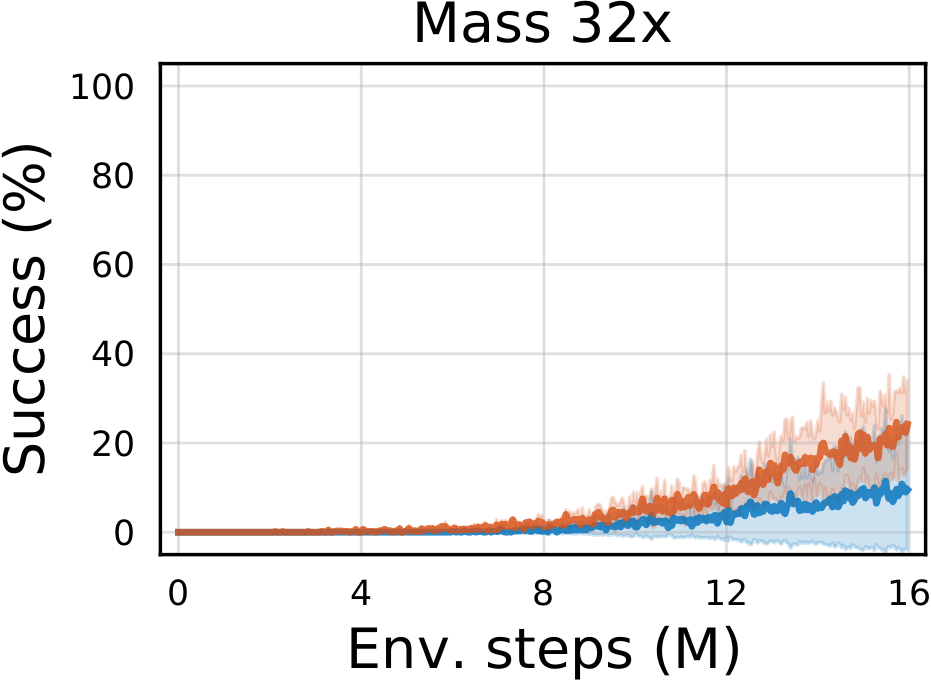}\\[1mm]
 \hspace{3mm}\includegraphicsgenim{width=0.41\linewidth}{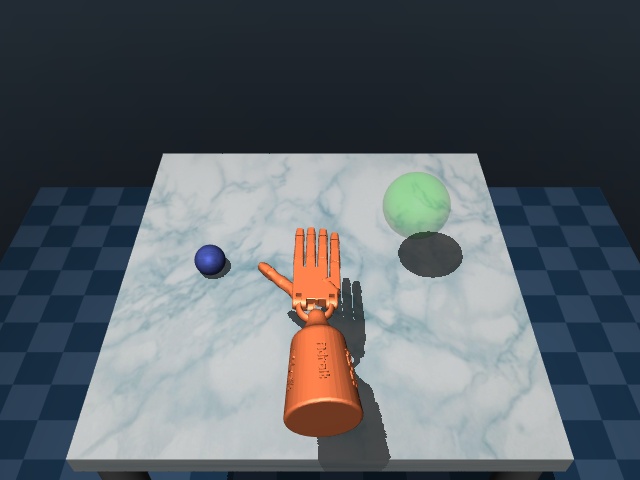}\hspace{1mm}
 \includegraphicsgenim{width=0.41\linewidth}{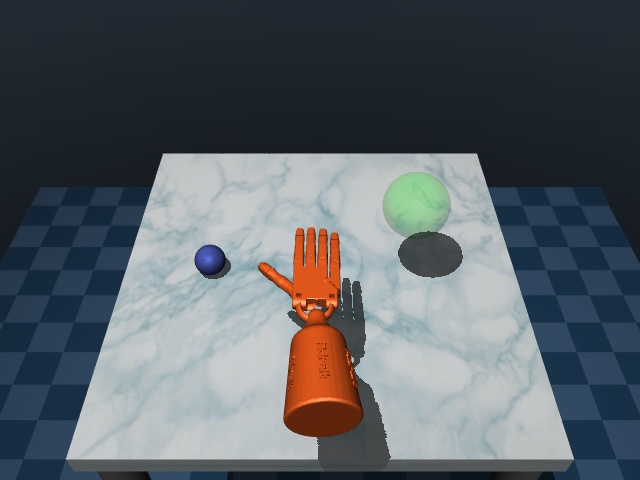}}}
\end{minipage}\hspace{0.2mm}
\begin{minipage}{.33\linewidth}\centering
 \subfloat[\textbf{Morphology.}\label{fig:gen_morph_fingers}]{\shortstack{
 \includegraphicstriml{20}{height=21.5mm}{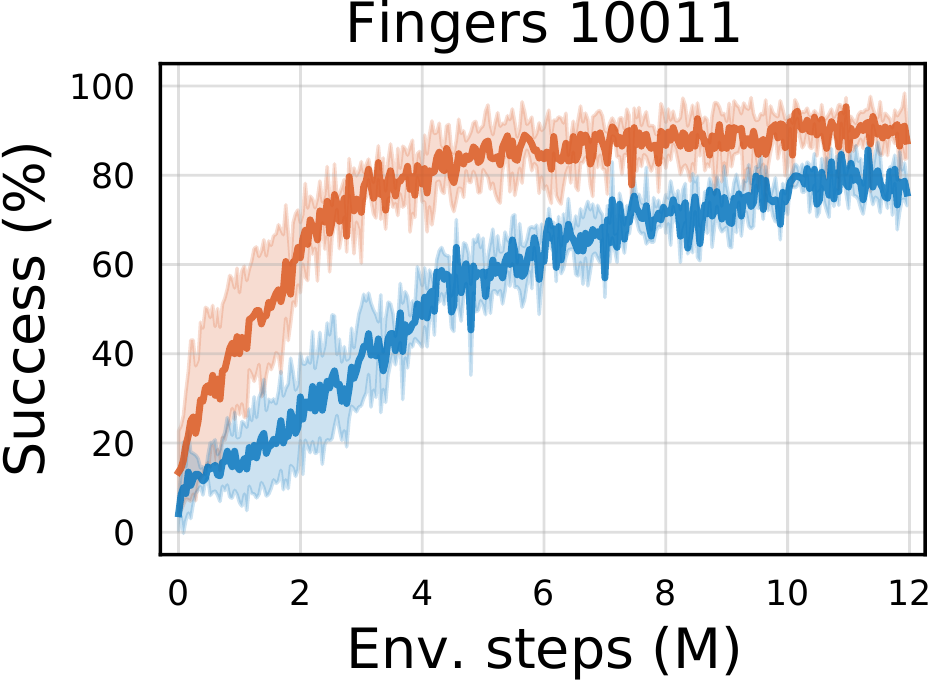}
 \includegraphicstriml{44}{height=21.5mm}{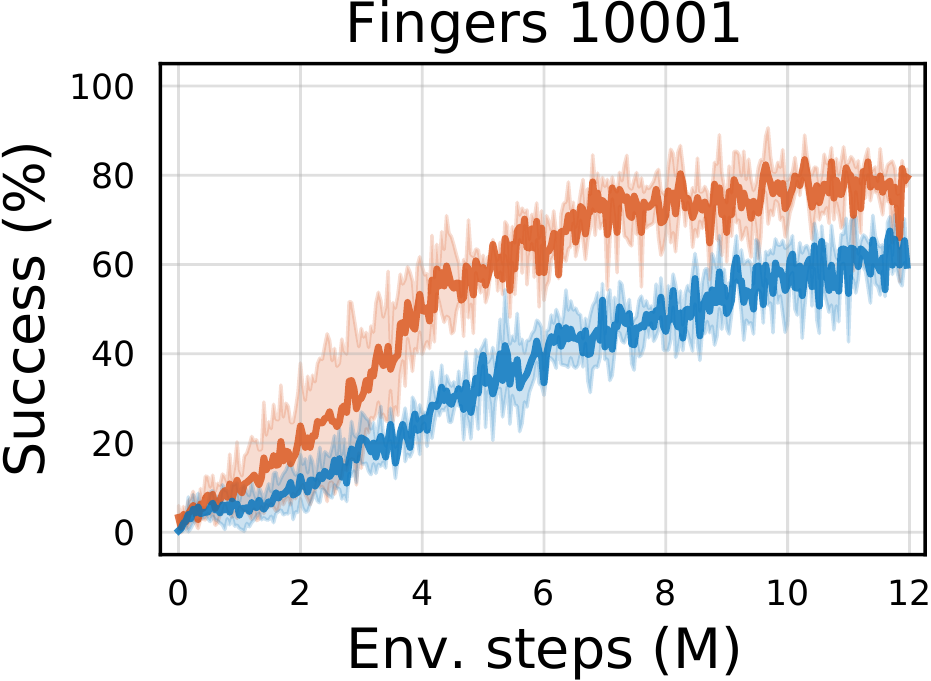}\\[1mm]
 \hspace{1mm}\includegraphicsgenim{width=0.41\linewidth}{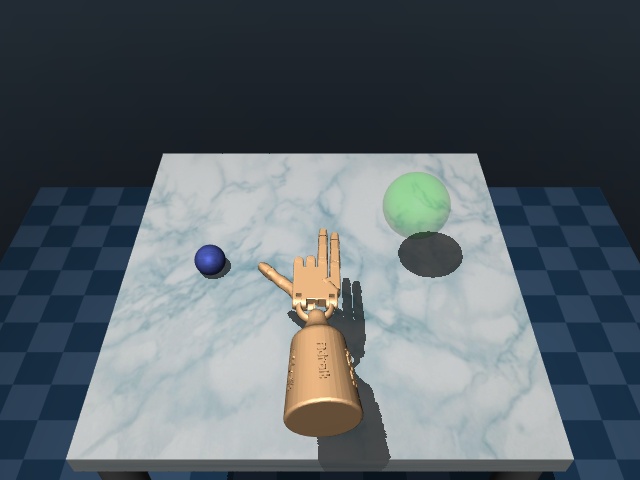}\hspace{1mm}
 \includegraphicsgenim{width=0.41\linewidth}{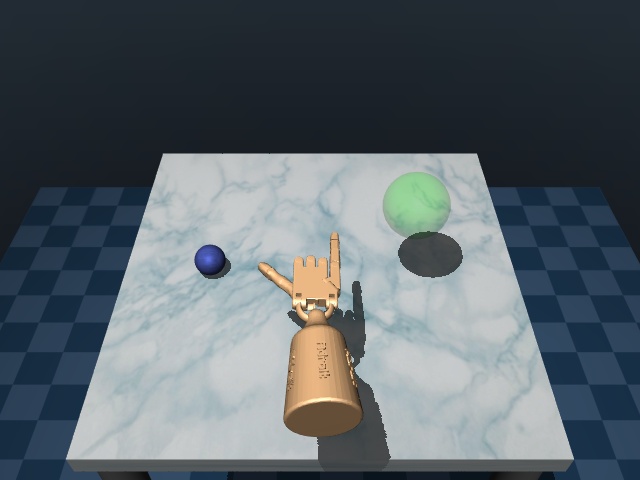}}}
\end{minipage}
\begin{minipage}{.33\linewidth}\centering
 \subfloat[\textbf{Objects.}\label{fig:gen_model_objects}]{\shortstack{
 \includegraphicstriml{20}{height=21.5mm}{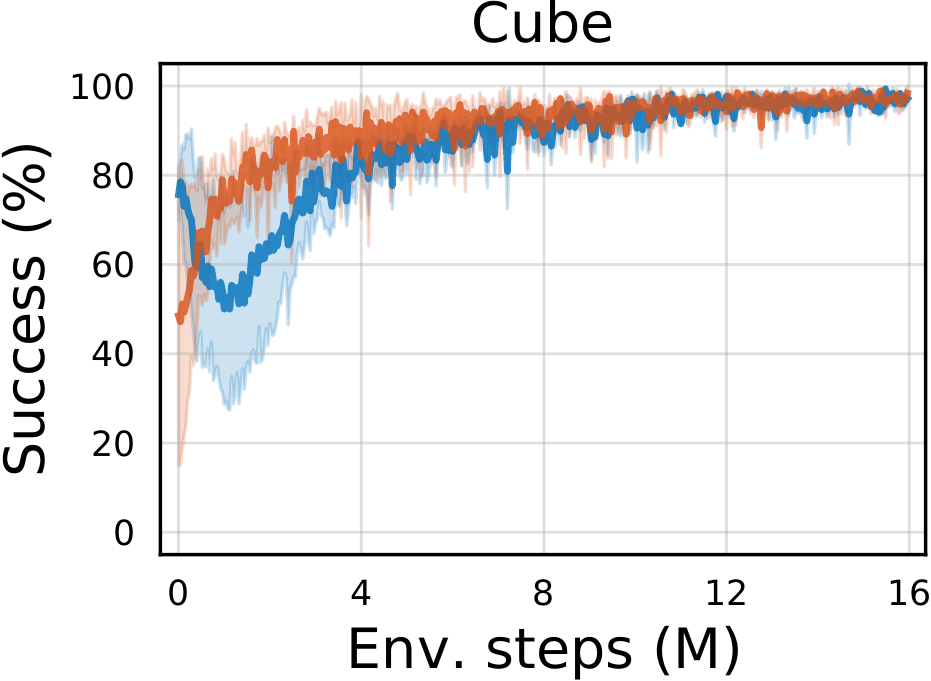}
 \includegraphicstriml{44}{height=21.5mm}{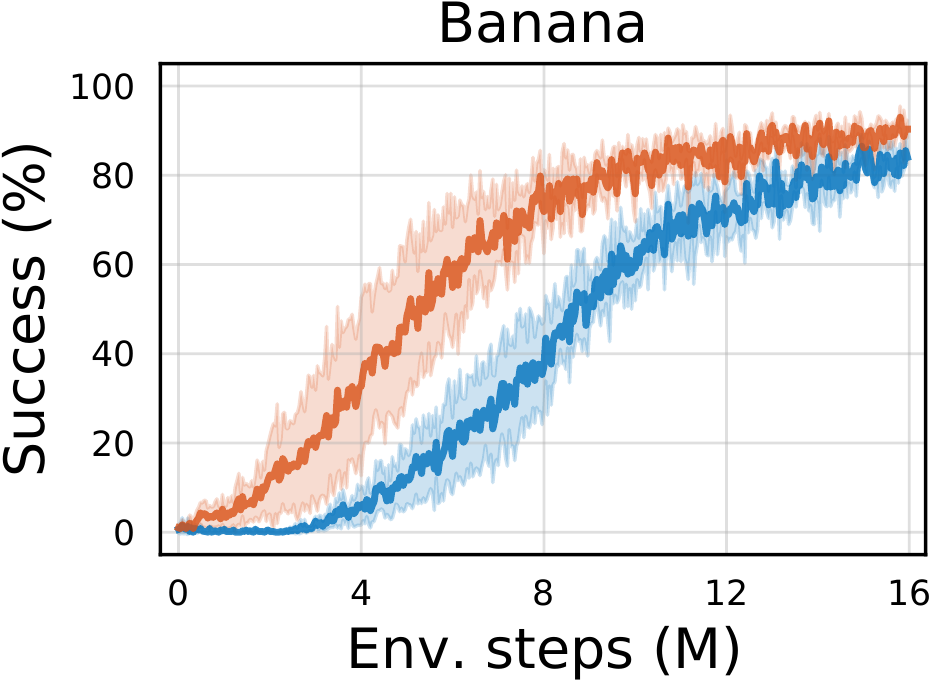}\\[1mm]
 \hspace{1mm}\includegraphicsgenim{width=0.41\linewidth}{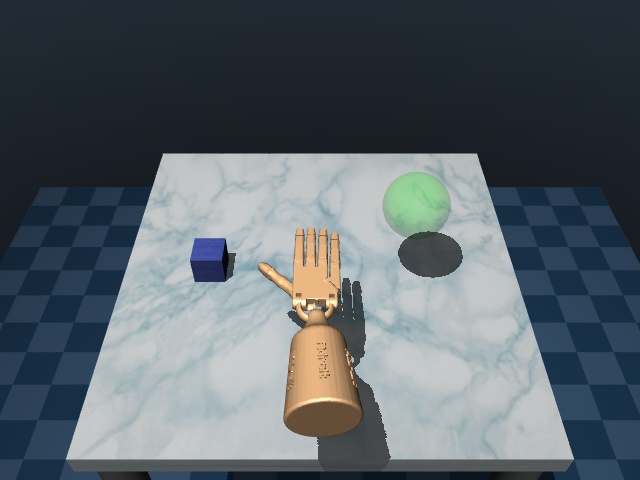}\hspace{1mm}
 \includegraphicsgenim{width=0.41\linewidth}{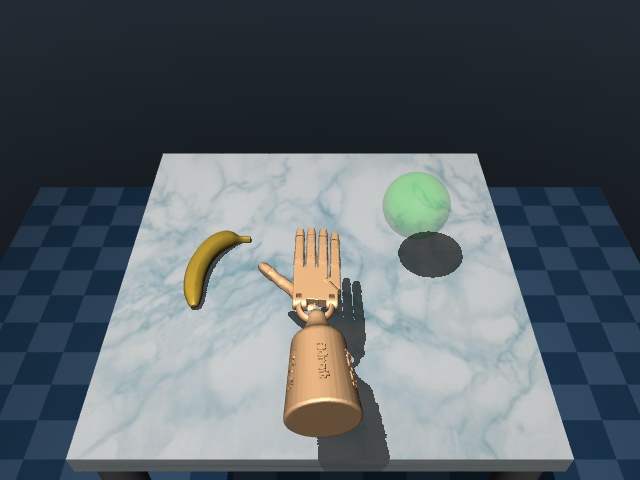}}}
\end{minipage}\vspace{-0mm}
\caption{\textbf{Demonstration mismatch.} We consider learning the object relocation task using \emph{original} demonstrations and \emph{different} dynamics, morphology, and objects. In all cases, we observe that our state-only method SOIL outperforms the state-action DAPG. This suggests that our method is able to adapt effectively and that state-only imitation may be preferable when there is a mismatch between the demonstrator and the learner settings.}
\label{fig:gen}\vspace{-2mm}
\end{figure*}

\subsection{State-Only Comparisons}\label{sec:state_only}

We now compare our method to different state-only imitation learning baselines.

\paragraph{Chamfer distance.} One way to leverage state-only demonstrations is to perform state matching between trajectories sampled from the current policy and demonstrations~\cite{Abbeel2010, Peng2018d}. We experiment with a number of variants based on standard distance functions including DTW~\cite{Needleman1970, Sakoe1978}, nearest neighbor, and Chamfer distance. We find the approach based on the Chamfer distance to work the best. Consequently, we adopt it as our state matching baseline in the comparisons.

\paragraph{Density estimation.} While state matching approaches can work well in certain settings, they rely on domain knowledge and careful tuning of the distance function, which limits their applicability in practice. Inspired by~\cite{Ho2016}, we explore a data-driven alternative. In particular, we train a density model to differentiate between states coming from the current policy and the demonstrations.

\paragraph{Comparisons.} In Figure~\ref{fig:comparison_state_only}, we compare our method SOIL to the two aforementioned baselines. First, we observe that both baselines work reasonably well on the easier pen in-hand manipulation task (right). Nevertheless, there is still a clear gap compared to SOIL. Second, we see that in case of the more challenging object relocation task SOIL outperforms the baselines considerably (left). This is a promising signal for the applicability of our method to harder settings.

\subsection{Ablation Studies}\label{sec:ablations}

Next, we perform ablation studies to get a better understanding of different aspects of our method.

\paragraph{Imitation term weights.} In Figure~\ref{fig:ablations_lams}, we study the impact of the imitation term weights from \Eqref{eq:soil}. First, we see that the scale of the imitation term plays an important role (left). In particular, using too small or too large of a weight leads to suboptimal performance. Thus, the scale needs to be tuned carefully. Second, we observe that annealing the imitation term to zero results in considerably better performance (right). This suggests that demonstrations are helpful during the initial exploration but start to hurt as the policy gets better and starts to exceed demonstrations.

\paragraph{Inverse dynamics model.} In Figure~\ref{fig:ablations_model}, we study the impact of the quality of the inverse dynamics model (left) on the policy (right). We observe that a good inverse dynamics model (low error) results in a good policy (high return), and vice versa. This suggests that the quality of the inverse dynamics model plays an important role in the effectiveness of our method.

\paragraph{Model-based RL enhancements.} Motivated by the advancements from the model-based RL literature, we explore inverse dynamics model training enhancements in Figure~\ref{fig:ablations_tricks}. First, we find that using a replay buffer reduces the variance (left). Consequently, we adopt the replay buffer in our method. Second, we see no clear benefit from using an ensemble of inverse dynamics models. Thus, we do not use an ensemble of inverse dynamics models in our method.

\subsection{Robustness to Demonstration Mismatch}\label{sec:generalization}

Finally, we evaluate robustness to a mismatch between the demonstrator and the learner settings.

\paragraph{Experimental setup.} Our setup is as follows. We study learning to perform the object relocation task using \emph{original} demonstrations and \emph{different} learning conditions. In particular, we consider different dynamics, morphologies, and objects. In these settings, using original actions may not be helpful and may even hurt. The motivation for these experiments is to test if our state-only method can be more robust to such settings than state-action approaches.

\paragraph{Different dynamics.} First, we consider different dynamics. Specifically, we perform training with a hand of increased mass. In this setting the model needs to learn a different action distribution. For example, in order to move a heavier hand the model needs to apply more force and in turn account for a larger momentum. In Figure~\ref{fig:gen_dynamics_mass}, we show the results for two variants of increasing difficulty (mass increased by 16$\x$ and 32$\x$). We observe a clear trend: our state-only method SOIL outperforms state-action based DAPG for different dynamics.

\paragraph{Different morphologies.} Next, we study different morphologies. In particular, we consider training to perform the ball relocation task using a hand with a subset of fingers. In this setting, the model must learn a different strategy compared to using the full hand. In Figure~\ref{fig:gen_morph_fingers}, we show the results for two hand configurations of varying difficulty with two to three fingers removed. In both cases, we observe that our state-only method performs better than DAPG.

\paragraph{Different objects.} Lastly, we consider different objects. For example, we study learning to relocate a banana from demonstrations of relocating a ball. In this setting, the grasping strategies suitable for different objects can vary considerably. In Figure~\ref{fig:gen_model_objects}, we report the results for two different object variants of increasing difficulty. In both cases, we observe that our state-only method SOIL outperforms DAPG that uses actions. Our gains are larger for the harder setting. Interestingly, even in the relatively easier case of the cube there is still a clear gap.

\section{CONCLUSION}

In this work, we explore state-only imitation learning for dexterous manipulation. To tackle this setting, we propose a simple and effective method that we call SOIL. The core idea is to train a policy and an inverse model jointly. Our method achieves results on par with state-action approaches and considerably outperforms RL without demonstrations. Going beyond, we show that SOIL can leverage demonstrations with different dynamics, morphologies, and objects. We hope that our work serves as a step toward the more general setting of learning from internet videos of humans. We encourage the readers to check the \href{https://people.eecs.berkeley.edu/~ilija/soil}{project page} for additional materials.

\clearpage
\bibliography{references}
\bibliographystyle{IEEEtran}

\end{document}